\documentclass[conference]{IEEEtran}

\usepackage{amsmath}
\usepackage{lipsum}
\usepackage{multicol}
\usepackage[pdftex]{graphicx}
\usepackage{graphicx}
\usepackage[export]{adjustbox}

\usepackage{float}
\floatstyle{plaintop}
\restylefloat{table}
\usepackage{placeins}
\usepackage{graphicx}
\usepackage{wrapfig}
\usepackage{color}
\usepackage{tabulary}
\usepackage{makecell}
\usepackage{array} 
\usepackage{amsmath}
\usepackage{amsthm}
\usepackage{algpseudocode}
\usepackage{algorithm}
\usepackage{wrapfig}
\usepackage{array, tabularx, caption, boldline}
\usepackage{graphicx}
\usepackage{cellspace}

\usepackage{setspace}
\usepackage{graphicx}
\usepackage{caption}

\usepackage{xcolor,colortbl}
\usepackage{url}
\usepackage{enumitem}
\usepackage{xcolor}
\usepackage{tabu}
\usepackage{setspace}
\usepackage[font={footnotesize}]{caption}
\usepackage{setspace}
\usepackage{float}
\usepackage{subfig}
\usepackage{multirow}
\usepackage{booktabs}

\begin{document}
	
	\title{ \huge \textbf{Representation Learning with Autoencoders for Electronic Health Records: A Comparative Study}}
	
	\author{\IEEEauthorblockN \centering {\quad \quad \quad Najibesadat Sadati$^{a}$ , Milad Zafar Nezhad$^{a}$, Ratna Babu Chinnam$^{a}$, Dongxiao Zhu$^{b,*}$}\\
		\IEEEauthorblockA{\quad \quad Department of Industrial and Systems Engineering,
			Wayne State University$^{a}$\\\quad \quad Department of Computer Science, Wayne State University$^{b}$\\
			\quad \quad Corresponding author$^{*}$, E-mail address: dzhu@wayne.edu}
	}
		
	\maketitle
	
\begin{abstract}
		
Increasing volume of Electronic Health Records (EHR) in recent years provides great opportunities for data scientists to collaborate on different aspects of healthcare research by applying advanced analytics to these EHR clinical data. A key requirement however is obtaining meaningful insights from high dimensional, sparse and complex clinical data. Data science approaches typically address this challenge by performing feature learning in order to build more reliable and informative feature representations from clinical data followed by supervised learning. In this paper, we propose a predictive modeling approach based on deep learning based feature representations and word embedding techniques. Our method uses different deep architectures (stacked sparse autoencoders, deep belief network, adversarial autoencoders and variational autoencoders) for feature representation in higher-level abstraction to obtain effective and robust features from EHRs, and then build prediction models on top of them. Our approach is particularly useful when the unlabeled data is abundant whereas labeled data is scarce. We investigate the performance of representation learning through a supervised learning approach. Our focus is to present a comparative study to evaluate the performance of different deep architectures through supervised learning and provide insights in the choice of deep feature representation techniques. Our experiments demonstrate that for small data sets, stacked sparse autoencoder demonstrates a superior generality performance in prediction due to sparsity regularization whereas variational autoencoders outperform the competing approaches for large data sets due to its capability of learning the representation distribution.\\ 

		
\end{abstract}

\IEEEpeerreviewmaketitle
\section{Introduction}
Healthcare is transitioning to a new paradigm under the emergence of large biomedical datasets in various domains of health. In fact, the explosive access to large Electronic Health Records (EHR) and advanced analytics is providing a great opportunity in recent years for improving the quality of healthcare \cite{li2017sdt}. The availability of patient-centric data brings about new opportunities in healthcare and enable data scientists to pursue new research avenues in the realm of personalized medicine using data-driven approaches.

Since EHRs are complex, sparse, heterogeneous and time dependent, leveraging them for personalized medicine is challenging and complicated to interpret. Representation learning (feature learning) provides an opportunity to overcome this problem by transforming medical features to a higher level abstraction, which can provide more robust features. On the other hand, labeling of clinical data is expensive, difficult and time consuming in general. However, there are many instances where unlabeled data may be abundant. Representation learning through unsupervised approaches is a very effective way to extract robust features from both labeled and unlabeled data and improve the performance of trained model based on labeled data.

There are several challenges in processing EHR data \cite{cheng2016risk}: 1) High-dimensionality, 2) Temporality which refers to the sequential nature of clinical events, 3) Sparsity, 4) Irregularity which refers to high variability, 5) Bias including systematic errors in the medical data, and 6) Mixed data types and missing data. Representation learning can overcome these challenges and the choice of data representation or feature representation plays a significant role in success of machine learning algorithms \cite{bengio2013representation}. For this reason, many efforts in developing machine learning algorithms focus on designing preprocessing mechanisms and data transformations for representation learning that would enable more efficient machine learning algorithms \cite{bengio2013representation}. There are several approaches for feature learning such as $K$-means clustering, Principal component analysis (PCA), Local linear embedding, Independent component analysis (ICA), and Deep learning. The key advantage of deep feature learning in comparison with the other shallow learning approaches is the potential to achieve more abstract features at higher levels of representation from non-linear transformation through multiple layers \cite{bengio2013representation}.

Deep representation learning \cite{bengio2013representation} includes a set of techniques that learn a feature via transformation of input data to a representation that can improve machine learning tasks such as classification and regression. In other words, representation learning helps to provide more useful information. These methods have shown great promise especially in the field of precision medicine and health informatics \cite{zhao2017learning}. Although deep representation learning has been explored in healthcare domain for predictive modeling and phenotyping, there is not a comprehensive comparison and evaluation of these techniques in real-world studies.

Deep learning models demonstrated promising performance and potential in computer vision, speech recognition and natural language processing tasks. The rising popularity of using deep learning in healthcare informatics is remarkable for several reasons. For instance, deep learning was recently employed to medicine and genomics to rebuilding brain circuits, performance prediction of drug molecules, identifying the effects of mutations on gene expressions, personalized prescriptions, treatment recommendations, and clinical trial recruitment \cite{miotto2017deep}. Deep learning is being employed in an unsupervised manner on EHRs for feature representation in order to achieve specific or general goals \cite{shickel2017deep}. For instance `Deep Patient' \cite{miotto2017deep} and `Doctor AI' \cite{choi2016doctor} approaches are good examples of these recent works which used unsupervised learning via deep learning before supervised learning.

In this research, we perform exploratory analysis to show the impact of deep feature representation on prediction performance and investigate the choice of deep learning approach across small and large datasets. For this purpose, we focus on two specific healthcare informatics problems using high dimensional EHR data. In the first case study, we use left ventricular mass indexed to body surface area (LVMI) to predict heart failure risk. In the second case study, we focus on two large and high dimensional datasets from eICU collaborative research database to predict patient length of stay (LOS) in ICU unit for two groups of patients.

\begin{figure*}
	\centering
	\includegraphics[scale= 0.48]{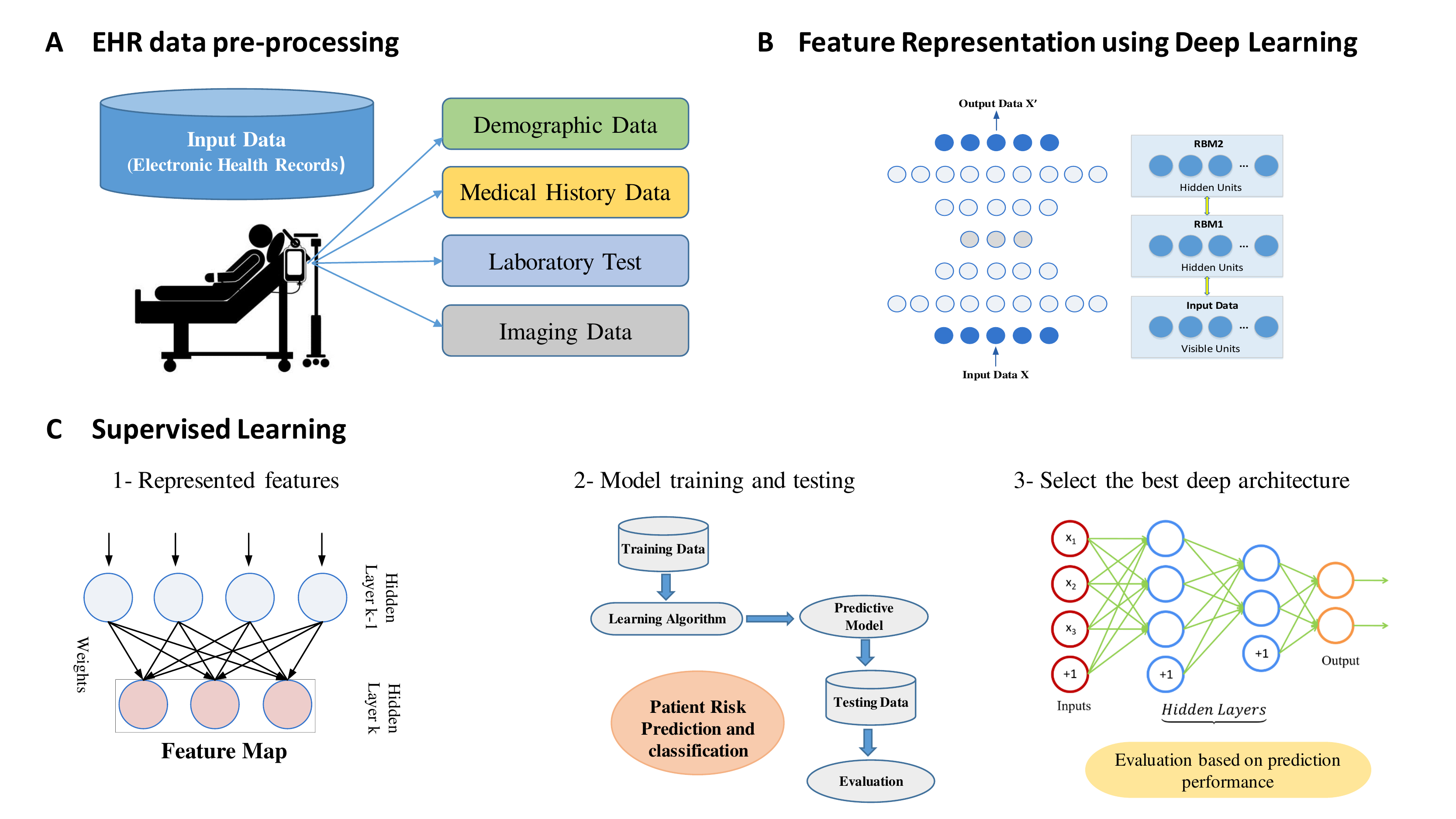}
	\caption{An illustration of the three consecutive steps of DIP approach}\label{Fig1}
\end{figure*}

Figure \ref{Fig1} demonstrates our deep integrated predictive (DIP) approach in three consecutive steps. First (step $A$), we start by preprocessing raw data to overcome some popular issues such as missing values, outliers and data quality. In the second step ($B$), we apply unsupervised deep learning for producing higher-level abstraction of input data. Finally (step $C$) we implement supervised learning for forecasting the target value and model evaluation. Based on the model evaluation results, steps $B$ and $C$ are applied iteratively to finalize and select the best deep architecture for feature learning.

Overall, we can label our approach as a semi-supervised learning framework where we apply the benefits of unsupervised learning to  different predictive tasks. While our framework utilizes deep learning for feature representation, our goal is to compare and evaluate the choice of deep network architectures through steps $B$ and $C$ as a back-and-forth approach among small and large EHR datasets. 

In general, we develop a predictive approach using deep learning and data representation for EHRs. In our method, we apply four deep architectures for feature representation in higher levels abstraction: Stacked sparse autoencoders, Deep belief network, Adversarial autoencoders and Variational autoencoders. Our contributions in this paper lie into two folds: 1) Improving predictive modeling by deep feature representation on EHRs where we apply various deep networks including generative autoencoders (AAE, VAE) and regular autoencoders (SSAE, DBN). 2) It is one of the first comparative studies to investigate the choice of deep representation among small and large datasets, and provide practical guidelines. This paper is an extension of our previous paper \cite{nezhad2016safs} with significant improvement and generalization.

The rest of the paper is organized as follows. Section 2 reviews the shallow feature learning methods and importance of deep feature learning approaches in health informatics. Section 3 explains the proposed prediction approach. Section 4 reports the EHRs data and implementation results. Finally, section 5 finishes with a discussion of results and conclusion.

\section{Related Works}
Since principal component analysis (PCA) is developed \cite{pearson1901liii}, feature learning (feature extraction or representation learning) has been researched for more than
a century to overcome the challenges of high dimensionality \cite{zhong2019shallow}. In this period, many shallow methods including linear and non-linear feature learning approaches have been developed until deep learning is introduced as an exciting new trend of machine learning in recent years. Deep learning demonstrated a great performance in feature representation through supervised and unsupervised learning approaches especially in the medical and health science domains. In this section, we first review the shallow feature learning methods and then we describe the importance of deep feature learning and review both supervised and unsupervised deep feature representation approaches.  


\subsection{Shallow Feature Learning}
The most popular shallow feature representation approach is Principle Component Analysis. PCA produces linear combinations of features, called principal components, which are orthogonal to each other, and can explain variation of features, and may achieve lower dimensionality. PCA is applied in several healthcare applications. Martis et al. \cite{martis2012application} used PCA for classification of ECG signals for automated diagnosis of cardiac health. Yeung et al. \cite{yeung2001validating} applied PCA to project gene expressions into lower dimension to cluster genes. In the other study \cite{ma2009identification}, authors proposed a systematic approach based on PCA to detect the differential gene pathways which are associated with the phenotypes.

In addition to standard PCA, several PCA-based approaches have been developed and applied in bioinformatics studies to improve the performance of PCA. For example, supervised PCA is proposed in \cite{bair2006prediction} in order to diagnose and treat cancer more accurately using DNA microarray data. Zou et al. \cite{zou2006sparse} proposed Sparse PCA using lasso (elastic net) regression for gene expression arrays. In the other research, Nyamundanda et al. \cite{nyamundanda2010probabilistic} applied Probabilistic PCA to analyze the structure of metabolomic data. 

Instead of PCA-based approaches, several other shallow feature learning methods have been developed with application to healthcare. Independent Component Analysis (ICA) \cite{yao2012independent} and Linear Discriminant Analysis (LDA) \cite{xu2009modified} are good examples of linear approaches. Among nonlinear dimensionality reduction, or so-called manifold learning approaches, isometric feature mapping (Isomap) \cite{li2010gene} and locally linear embedding (LLE) \cite{tenenbaum2000global} are the most popular models. 

The key benefit of using shallow feature learning is the capability to interpret the represented features. Shallow feature learning approaches are efficient computationally and the learning process is straight forward but they have not demonstrated great performance in high dimensional and complex data such as temporal/spatial data or image data because they cannot be stacked to provide deeper and more abstract representations \cite{bengio2013representation}.

\subsection{Deep Feature Learning}
Deep feature representation has been employed in several areas using EHRs (e.g. diagnosis and medication data), genomics data, medical text and imaging data with various purposes including risk factors selection, disease phenotyping and disease risks prediction or classification \cite{miotto2017deep}. Feature representation using deep learning can be achieved either by supervised deep learning predictive models (e.g., deep feed-forward neural network and convolutional nets) or by unsupervised deep learning approaches (e.g., deep autoencoders). In this section, we review the related studies for both approaches in biomedical and healthcare applications. 

\subsubsection{Deep Supervised Feature Learning}
The supervised deep predictive approaches extract features through learning weights and biases with considering target variable in the cost function. As a good example, Li et al. \cite{li2016deep} proposed a novel deep feature selection model for selecting significant features inputted in a deep neural network for multi-label data. The authors used elastic net regularization to select most important features. They added a one-to-one linear layer between the visible layer and the first hidden layer of a multi-layer perception (MLP) to rank features based on regularized weights in the input layer obtained after training. Finally, they applied their model to a genomics dataset. In another study, Cheng et al. \cite{cheng2016risk} first represented the EHRs as a temporal matrix with two dimensions, time and event for all records and then used four-layer CNN for extracting phenotypes and applying prediction for two case studies: congestive heart failure and chronic obstructive pulmonary disease. 

Choi et al. \cite{choi2016doctor} proposed a predictive framework termed Doctor AI for medical events. The authors employed a recurrent neural network (RNN) on large-scale temporal EHR data to predict the diagnosis and medication categories for further visits. Zhao et al. \cite{zhao2016multiscale} developed a brain tumor image segmentation method using convolutional neural networks (CNNs). They applied their approach on multimodal brain tumor image segmentation benchmark (BRATS) data and obtained advanced accuracy and robustness. 

Feature learning through deep supervised predictive approach requires large scale labeled data for training while in many healthcare applications it is hard to collect enough labeled data. Unsupervised and semi-supervised feature learning approaches can overcome the label scarcity problem and provide better feature representation.

\subsubsection{Deep Unsupervised Feature Learning}
Many studies used unsupervised or semi-supervised deep feature learning for EHRs and applied predictive models on top of represented features. Miotto et al. \cite{miotto2016deep} applied stack denoising autoencoders (SDA) for feature learning and representation of large scale electronic health records. They used EHRs of approximately $700,000$ individuals related to several diseases including schizophrenia, diabetes, and various cancers. Their model improved medical prediction, which could offer a machine learning framework for augmenting clinical decision systems. 
Recently, Che at al. \cite{che2017boosting} proposed a semi-supervised framework for EHRs risk prediction and classification. They developed a modified  generative adversarial network called ehrGAN for feature representation and used CNN for performing prediction task. In the other research,  \cite{nezhad2016safs}, authors proposed a novel feature selection model using deep stacked autoencoders. They performed their approach on a health informatics problem to identify the most important risk factors related to African-Americans who are in risk of heart failure. Wulsin et al. \cite{wulsin2010semi} developed an approach using deep belief nets for electroencephalography (EEG) anomaly detection to monitor brain function in critically ill patients.\\
Cao et al \cite{cao2016deepqa} trained deep belief network on several large datasets for prediction of protein tertiary structure for protein quality assessment based on different perspectives, such as physio-chemical and structural attributes. In the other study \cite{deng2019collaborative}, a healthcare recommender system developed based on variational autoencoders and collaborative filtering. Authors used VAE to learn better relationships between items and users in collaborative filtering. \\
Our predictive approach for cardiovascular risk level (LVMI) and length of stay (LOS) in ICUs can be considered in the last group. Readers for more comprehensive review about deep feature learning applications in healthcare can refer to recent review articles provided by Ravi et al. \cite{ravi2017deep}, Litjens et al. \cite{litjens2017survey}, Miotto et al. \cite{miotto2017deep}, Xiao et al. \cite{xiao2018opportunities} Shickel et al. \cite{shickel2018deep} and Purushotham et al. \cite{purushotham2018benchmarking}.\\
Although deep feature learning (supervised, unsupervised and semi-supervised) generally provides better representation rather than shallow approaches, but it is hard to interpret and computationally expensive to train with several hyperparameters. Therefore, the right choice of deep network can reduce significant effort in training process. In this way, we try to provide empirical insights for choice of deep representation approach across small and large datasets which has not been studied in the literature.

\section{Methodology}
\begin{figure}[H]
	\centering
	\includegraphics[scale= 0.55]{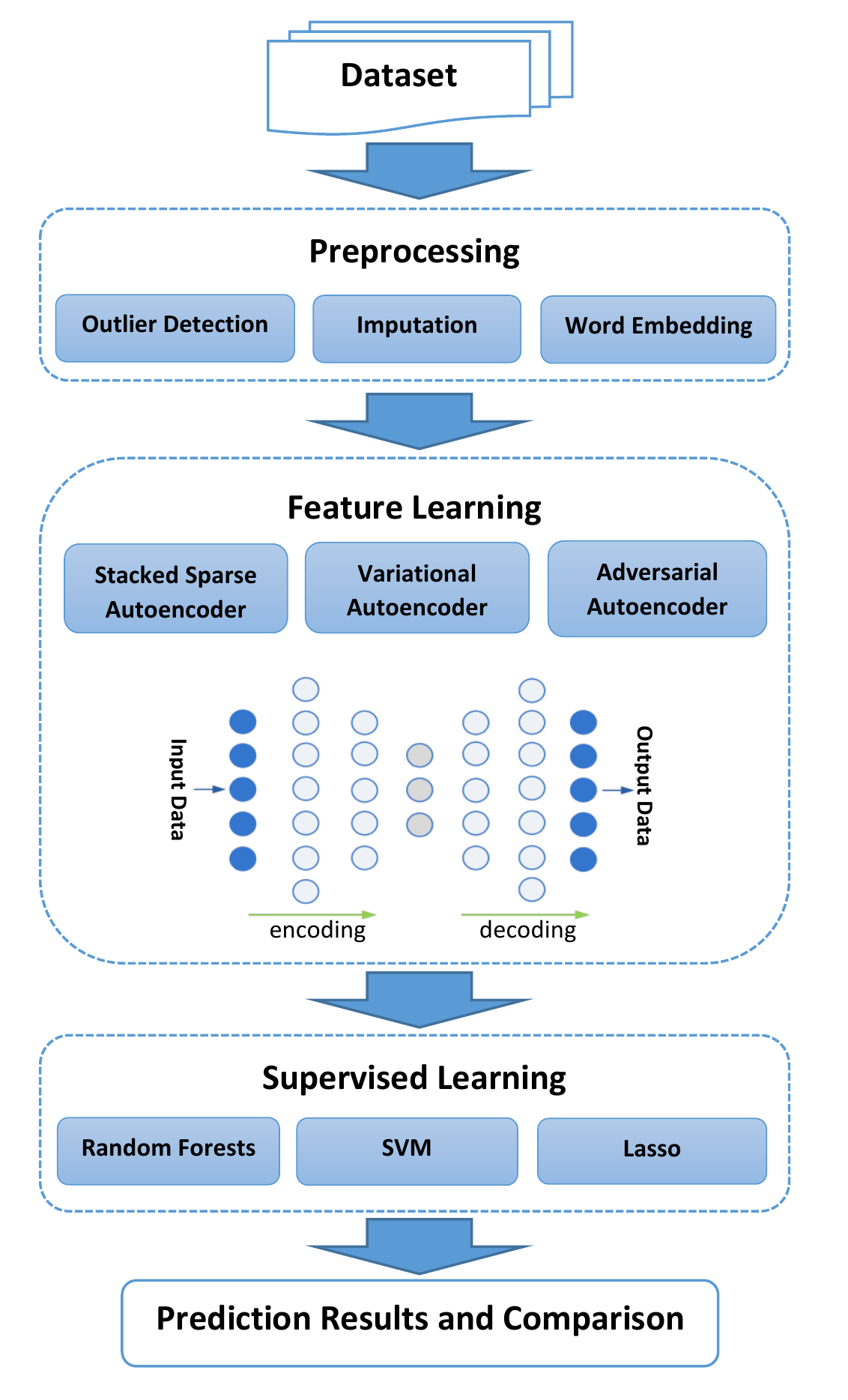}  
	\caption{The technical workflow of the proposed approach}
	\label{figure1}
\end{figure}

In this study, we propose a comprehensive evaluation study using an integrated predictive framework for deep representation learning. We use this framework to compare different deep networks to solve healthcare informatics problems in predictive modeling. Our methodology follows the work flow shown in Figure \ref{figure1} that includes three consecutive steps.

\subsection{Step$-$1: Preprocessing and Word Embedding} 
In the first step, we use the preprocessing methods such as outlier detection and imputation for missing values existed in the dataset. We also transform categorical and text variables to vectors using a well-known word embedding algorithm by applying "text2vec" package in R. Discovering efficient representations of discrete categorical and text features have been a key challenge in a variety of biomedical and healthcare applications \cite{choi2016multi}. Word Embedding algorithms are developed to map the categorical and text features (words) to vectors of real numbers. Among several approaches for word embedding in the literature such as Matrix Factorization methods and Shallow Window-Based methods, we use Glove algorithm \cite{pennington2014glove} as a well-known algorithm for word representation. GloVe algorithm uses the global word co-occurrence matrix to learn the word representations.

\subsection{Step$-$2: Feature Representation using Deep Learning} 
In the second step, all features will be represented in higher-level abstraction by four different deep autoencoder networks separately: 1) Stacked sparse Autoencoder (SSAE), 2) Deep Belief Network (DBN), 3) Adversarial Autoencoder (AAE) and 4) Variational Autoencoder (VAE). The performance of each network could be various in the different cases and it is necessary to consider hyper-parameters tuning such as learning rate, batch size, number of epochs, and number of hidden layers and hidden units for precise training to avoid over-fitting. The four deep learning architectures are explained as follows:

\subsubsection{Stacked Sparse Autoencoder} 
\begin{figure}[H]
	\centering
	\includegraphics[scale= 0.30]{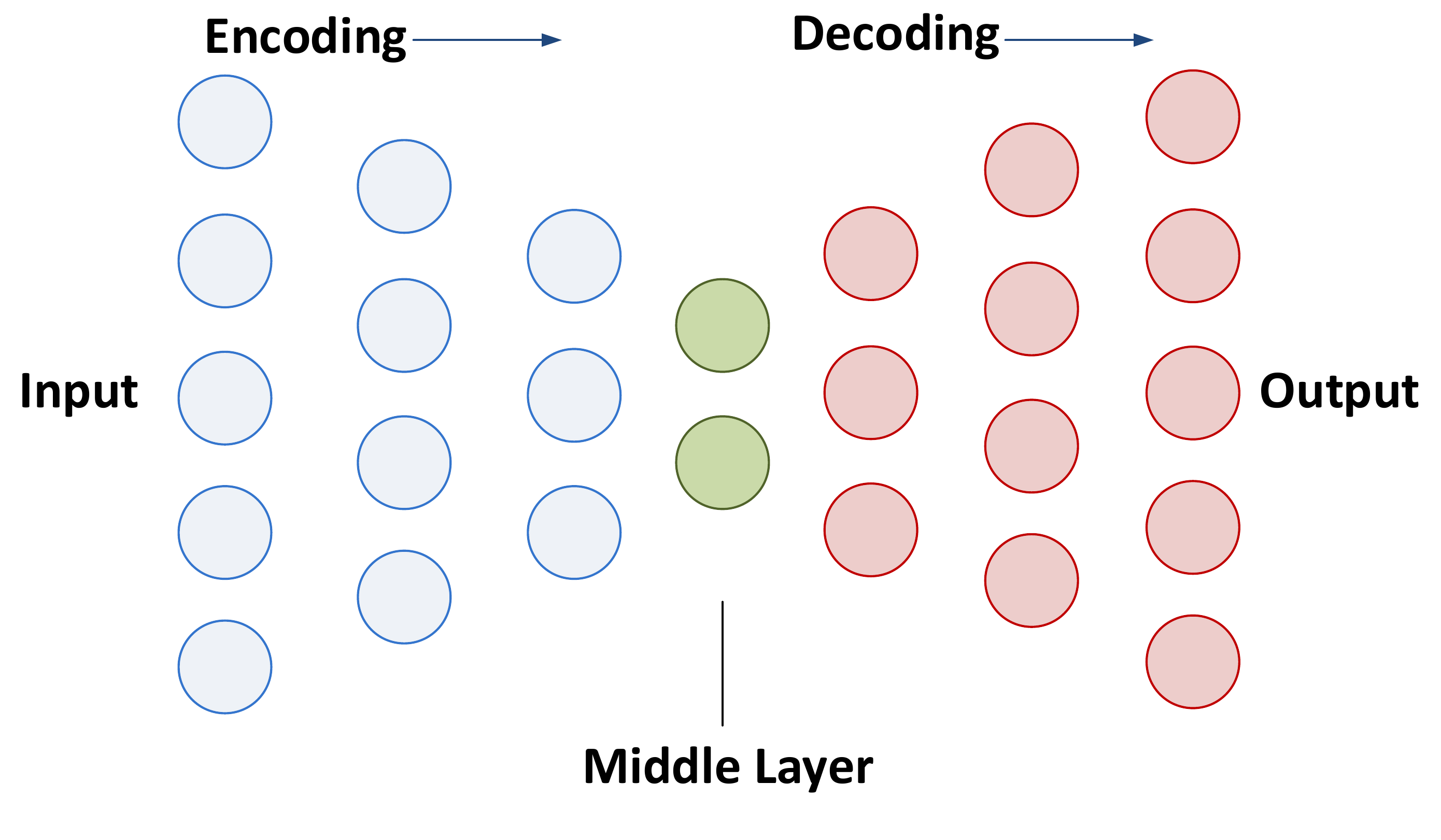}  
	\caption{Stacked Autoencoders (SAE) architecture}
	\label{fig_fa}
\end{figure}

An autoencoder network is an unsupervised learning methodology which number of output layer's neurons are equal to the number of input layer's neurons. AE tries to reconstruct input data ($x$) in output ($x^{'}$) layer by encoding and decoding process \cite{hinton2006reducing, vincent2010stacked}. AEs consist of an encoder, which converts the input to a latent representation, and a decoder, that remodels the input from this representation. Autoencoders are trained to minimize the reconstruction errors. Stacked Autoencoders is a deep network consisting of multiple layers of autoencoders (Figure \ref{fig_fa}) which can be trained in layer wised approach \cite{bengio2007greedy}. After training of deep network, middle layer illustrates the highest-level representation of original features \cite{learning2013computer}.

The loss function for training an autoencoder can be defined as following:
\begin{align}
&Loss(x, x^{'})= \Vert x- x^{'}\Vert = \Vert x- f(W^{'} (f(Wx+b))+b^{'}) \Vert ,&
\end{align}
where $f$ is the activation function and $W$, $W^{'}$, $b$ and $b^{'}$ are the parameters of the hidden layers. 

The above loss function is reliable when the number of hidden units in the latent layer being small, but even in the case of large hidden units (even greater than the number of input features), which is called sparse representation or stacked sparse autoencoder (SSAE), we can still explore reliable architecture, by imposing sparsity constraints on the network \cite{ng2011sparse}. In Sparse autoencoders, we formulate the loss function with regularizing activations (not weights of the network) and we encourage the learners to train encoding and decoding based on activating a small number of neurons. We can impose this sparsity constraint by adding $L1$ regularization or KL-Divergence (E.q \ref{sparse1}) to the loss function \cite{ng2011sparse}. 
\begin{align}
&Loss(x, x^{'})+ \lambda \sum_{i}\mid a_{i}^{(h)}\mid \quad or \quad Loss(x, x^{'})+ \lambda \sum_{j} KL(\rho \Vert \bar \rho_{j} ).  &
\label{sparse1}
\end{align}

\subsubsection{Deep Belief Network} 
\begin{figure}[H]
	\centering
	\includegraphics[scale= 0.33]{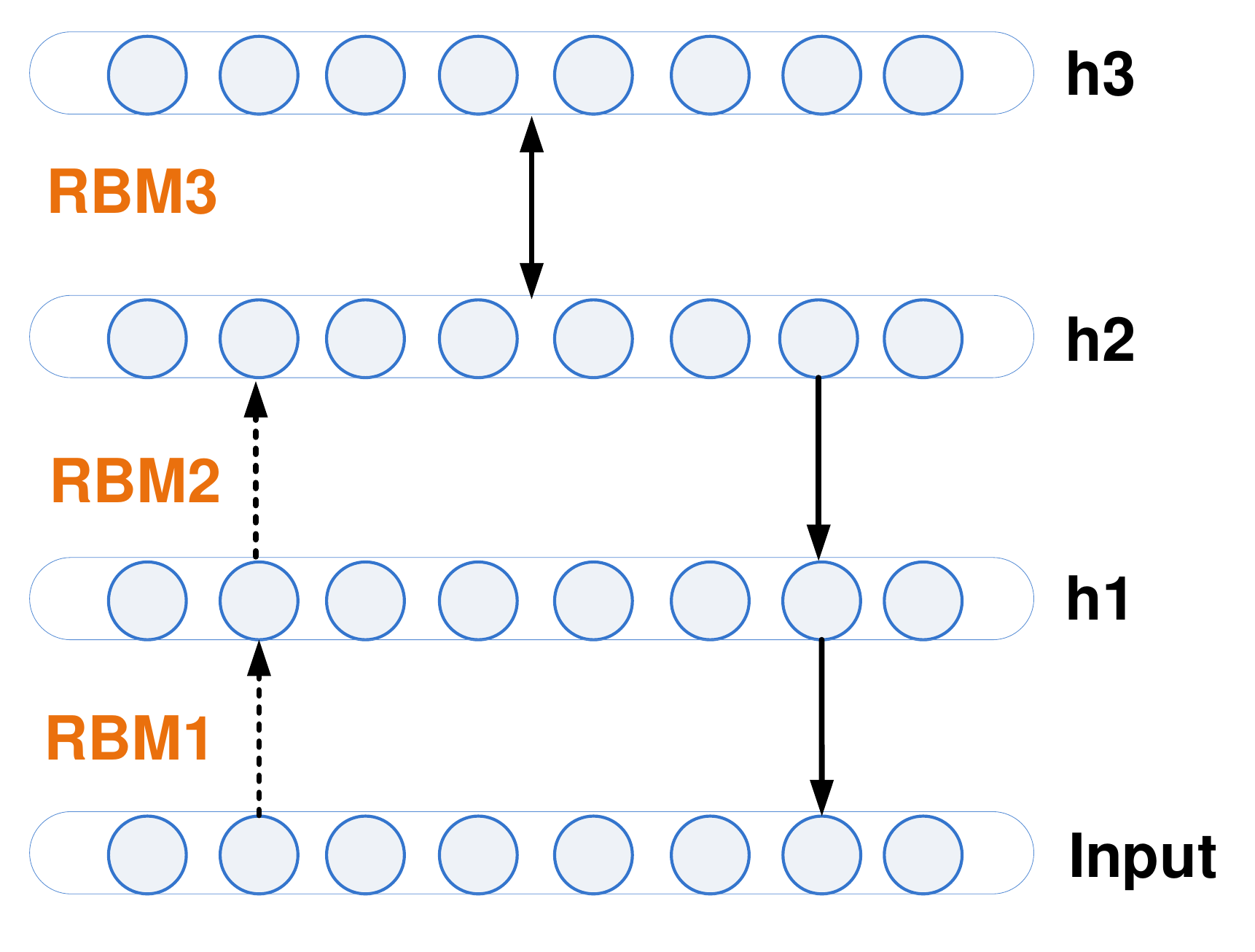}  
	\caption{Deep Belief Network (DBN) architecture}
	\label{fig_fb}
\end{figure}

Deep Belief Networks are graphical models that are constructed by stacking of several RBMs to get better performance rather than individual RBM. Hinton and Salakhutdinov \cite{hinton2006reducing} showed that DBNs can be trained in greedy layer-wise unsupervised learning approach. They defined the joint probability distribution between visible and hidden layers as follows: 
\begin{align}
& P(x, h^{1}, ..., h^{l}) =  \prod_{k=0} ^{l-2}  P(h^{k}|h^{k+1}) P(h^{l-1} h^{l}),&
\end{align}
where, $x$= $h^{0}$, $P(h^{k-1}|h^{k})$ is a conditional distribution for the visible units conditioned on the hidden units of the RBM at level $k$, and $P(h^{l-1}, h^{l})$  is the visible-hidden joint distribution in the top-level RBM. This is illustrated in Figure \ref{fig_fb}.


In the layer-wised training, the input layer (visible unit) is trained as a RBM and transformed into the hidden layer based on optimizing of log-likelihood as below \cite{hinton2006fast}:
\begin{align}
& \log p(x) = KL(Q(h^{(1)}|x)\Vert p(h^{(1)}|x)) + H_{Q(h^{(1)}|x)} +&\\\notag 
&\sum_{h} Q(h^{(1)}|x)(\log p(h^{(1)})) + \log p(h^{(1)}).&
\end{align}

$KL(Q(h^{(1)}|x)\Vert p(h^{(1)}|x))$ is the KL divergence between $Q(h^{(1)}|x)$ of the first RBM and $p(h^{(1)}|x)$. Then the first layer's represented hidden units will be considered as input data (visible units) for the second layer and this process continues. Readers for more detail about the training process can refer to \cite{hinton2006fast} and \cite{bengio2007greedy}.

\subsubsection{Variational Autoencoder} 
Variational Autoencoder (VAEs) is one of the most popular approaches to representation learning developed in recent years. Variational autoencoders are probabilistic generative models and have the same architecture as autoencoders, but consider specific assumptions about the distribution of middle/latent layer variables. Variational autoencoders learn the true distribution of input features from latent variables distribution using Bayesian approach and present a theoretical framework for the reconstruction and regularization purposes \cite{tabacof2016adversarial}:
\begin{align}
&p(x) = \int p(x,z)dz = \int p(x|z)p(z)dz.	&
\label{Eqvar}
\end{align}

In Eq. (\ref{Eqvar}), $p(x|z)$ is the probability function of the observed data and the output of the decoder network by considering noise terms. In this equation, $z$ is the latent representation and $p(z)$ is the representation prior with an arbitrary distribution such as standard normal distribution or a discrete distribution like as Bernoulli distribution. There exist two problems for solving above equation: defining the latent variables $(z)$ and marginalizing over $z$. The key intention behind the variational autoencoder is to try to sample values of $z$ that are likely to have generated $x$ and compute $p(x)$ for these values. To make tractable above integral, an approach is to maximize its variational lower bound using the Kullback-Leibler divergence (KL divergence or D) as follows: \cite{doersch2016tutorial}:
\begin{align}
&\text{\normalsize $E_{q_{\phi}(z|x)}[\log^{p(x|z)}]-D(q(z|x) \parallel p(z)) =$}&
\notag\\ 
&\text{\normalsize$log^{p(x)} - D[q(z|x) \parallel p(z|x)]$ }.& 
\label{Eqvar2}
\end{align}	
We can apply Bayes rule to $p(z|x)$ and reformulate Eq. (\ref{Eqvar2}):
\begin{align}
&log^{p(x)} - D[q(z|x) \parallel P(z|x)] = &\notag\\ 
&E_{z\sim q} [log^{p(x|z)}] - D [q(z|x) \parallel p(z)],  &
\label{Eqvar3}
\end{align}
while $q$ is encoding $x$ to $z$ and $p$ is decoding $z$ to reconstruct input $x$. The structure of variational autoencoder has been illustrated in Figure \ref{fig_f1}. 

\subsubsection{Adversarial Autoencoder} 
Adversarial autoencoder (AAE) is a probabilistic autoencoder based on generative adversarial networks (GAN) \cite{goodfellow2014generative} which propose a minmax game among two neural network models: generative model ($G$) and discriminative model ($D$). The discriminator model, $D(x)$, is a neural network that estimate the probability of a point $x$ in data space came from data distribution (true distribution which our model is training to learn) rather than coming from generative model \cite{makhzani2015adversarial}. At the same time, the generator model, $G(z)$, tries to map sample points $z$ from the prior distribution $p(z)$ to the data space. $G(z)$ is trained by maximum confusing of discriminator in trusting that samples it produces; originated from the data distribution. The generator is trained by using the gradient of $D(x)$ related to $x$, and using that to improve its parameters. The solution of this game can be represented as below:
\begin{align}
&\min_G \max_D E_{x\sim p_{data}} [\log D(x)] + E_{z\sim p(z)} [\log (1- D(G(z)))].  &
\end{align}	

\begin{figure*}[h!t]
	\centering
	\subfloat[VAE architecture 
	]{\includegraphics[width=0.45\textwidth]{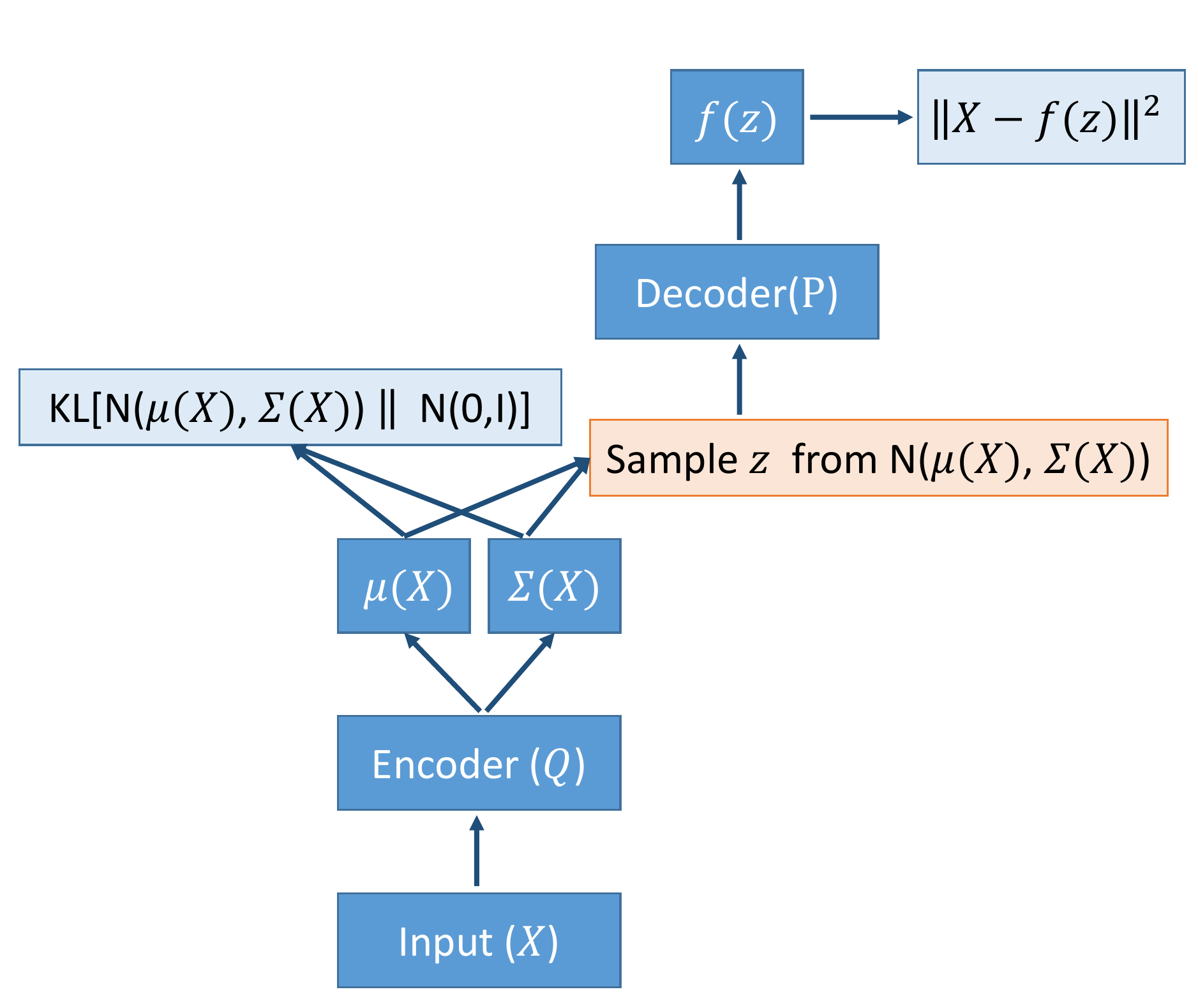}\label{fig_f1}}
	\hfill
	\subfloat[AAE architecture ]{\includegraphics[width=0.45\textwidth]{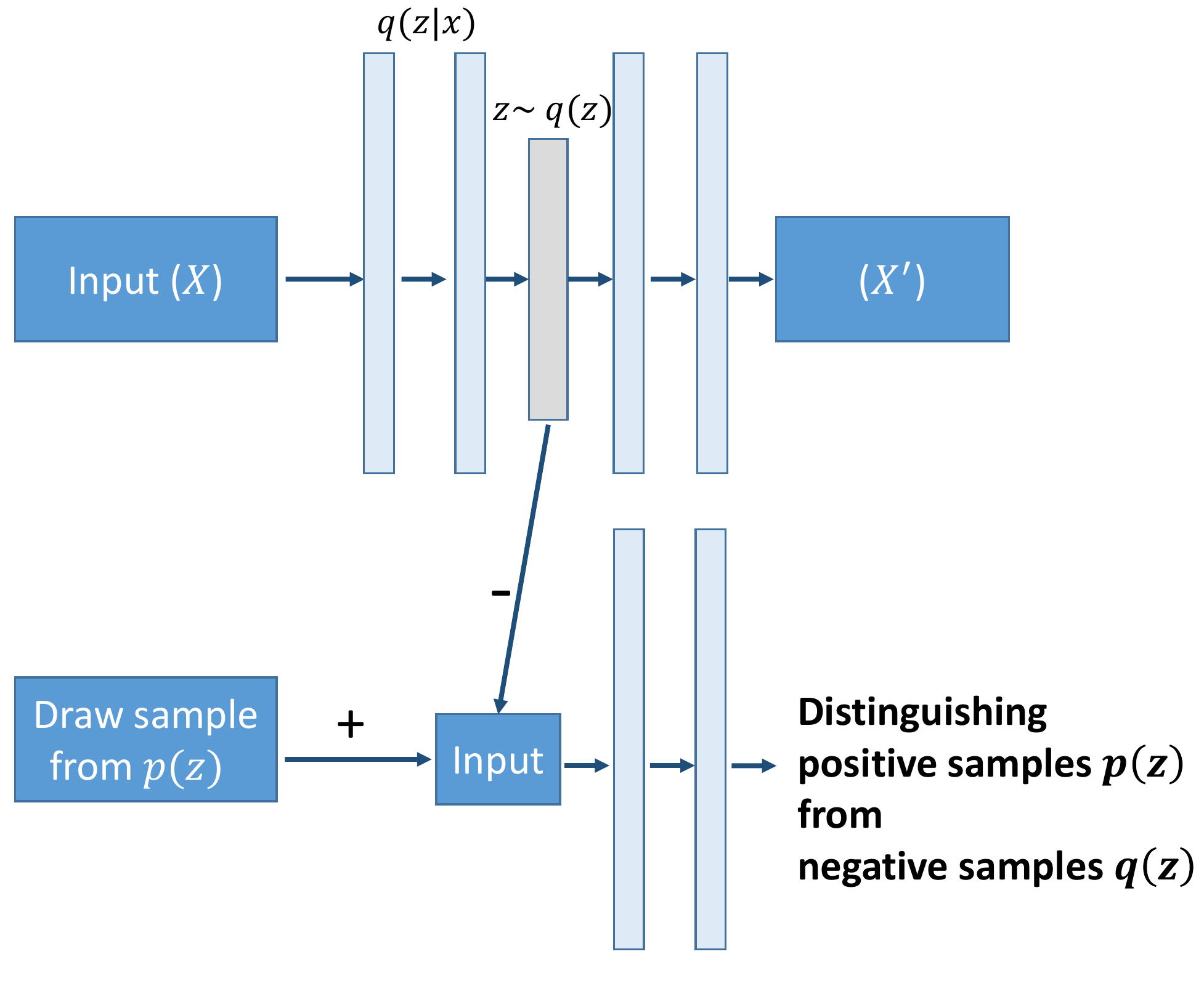}\label{fig_f2}}
	\caption{(a) Variational autoencoder network, where $P(X|z)$ is Gaussian distribution, (b) Adversarial autoencoder network, where the top row is a standard autoencoder and the bottom row shows a second network trained to discriminatively classify whether a sample arises from the latent layer or from a arbitrary distribution.}
	\label{aae_vae}
\end{figure*}

The adversarial autoencoder uses similar idea of GAN in training true distribution of data space by matching the aggregated posterior of latent variables to an arbitrary prior distribution in the reconstruction and the regularization phases \cite{makhzani2015adversarial}. In the other word, The adversarial autoencoder is an autoencoder that is regularized by coordinating the aggregated posterior, q(z), to an arbitrary prior, p(z). Simultaneously, the autoencoder attempts to minimize the reconstruction error. The architecture of an adversarial autiencoder is shown in Figure \ref{fig_f2}. 

The summary of deep networks applied in this study has been described in table \ref{summary}.

\begin{table*}[t]
	\small 
	\centering
	\captionsetup{justification=justified, width=0.9\linewidth}
	\caption{Summary and comparison of deep networks used in this study } \label{summary}
	
	\begin{tabular}{ p{7 cm}  p{7 cm}  }
		\midrule[1.5pt]
		\textbf{ Architectures Description} & \textbf{Key Points}\\
		\midrule[1.5pt]
		\textbf{1. Stacked Sparse Autoencoder} \begin{itemize}
			\item Proposed in \cite{hinton2006reducing, ng2011sparse} with the goal of dimensionality reduction
			\item AE tries to reconstruct input data in output layer by encoding and decoding process 
		\end{itemize}  & 	
		\textbf{Pros} \begin{itemize}			
			\item Sparse autoencoder is appropriate for small data through regularization.
			\item  Sparse autoencoder can be fine tuned easily by itself using ordinary back-propagation approach
		\end{itemize}  
		\textbf{Cons} \begin{itemize}
			\item The pre-training step is needed
			\item Vanishing errors may cause problem in training step
		\end{itemize} \\
		\midrule[0.5pt]
		
		\textbf{2. Deep Belief Network} \begin{itemize}
			\item Introduced in \cite{hinton2006fast} constructed by stacking of several RBMs
			\item DBNs are graphical models that can be trained based on greedy-layer wised approach
			\item Only the connection  between top layers is undirected
		\end{itemize}  & 	
		\textbf{Pros} \begin{itemize}
			\item Take the advantages of energy-based loss function instead of ordinary one
			
		\end{itemize}  
		\textbf{Cons} \begin{itemize}
			\item Training process is computationally expensive 
			\item  Fine tuning of DBNs seems to be difficult
		\end{itemize}  \\
		\midrule[0.5pt]
		
		\textbf{3. Variational Autoencoder} \begin{itemize}
			\item Proposed in \cite{kingma2013auto} to learn the true distribution of input features from latent space distribution using Bayesian approach
			\item VAEs apply a KL divergence term to impose a prior on the latent layer
			
		\end{itemize}  & 	
		\textbf{Pros} \begin{itemize}
			\item VAEs are flexible generative model
			\item VAE is a principled approach to generative models
			
		\end{itemize}  
		\textbf{Cons} \begin{itemize}
			\item Approximation of true posterior is limited 
			\item  VAEs Can have high variance gradients
		\end{itemize}  \\
		\midrule[0.5pt]
		
		\textbf{4. Adversarial Autoencoder} \begin{itemize}
			\item Proposed in \cite{makhzani2015adversarial} to impose the structure of input data on the latent layer of an autoencoder
			\item Adversarial autoencoders are generative autoencoders that use adversarial
			training to match the distribution of an arbitrary prior on the latent space
			
		\end{itemize}  & 	
		\textbf{Pros} \begin{itemize}
			\item Flexible representation to impose arbitrary distributions on the latent layer.
			\item It can capture any distribution for generation sample, both continuous and discrete
			
		\end{itemize}  
		\textbf{Cons} \begin{itemize}
			\item It is challengeable to train because of the GAN objective
			\item  It is not scalable to higher number of latent variables
		\end{itemize}  \\
		\midrule[1.5pt]
		
	\end{tabular}
\end{table*}

\subsection{Step$-$3: Supervised Learning} 
In this step, we apply supervised learning models on the top of represented dataset for all four feature extraction approaches. Once the features are extracted, these representations from main dataset are entered in a linear and non-linear supervised regression models such as Random Forests, SVM and LASSO for prediction. Finally we evaluate and compare the performance of feature learning step by RMSE based on the prediction models. 

\section{Experimental Study}
In our experimental study, we implement our methodology on three different EHRs datasets. First, we use a small dataset related to cardiovascular disease with high dimensional features, then we apply our method on two large datasets from eICU collaborative research database. This study design (considering small and large datasets) helps us to discover the performance of our method in different scenarios and compare the choice of representation learning for each one. 

\subsection{Case study 1 (Small Dataset): DMC dataset}
Cardiovascular disease (CVD) is the leading cause of death in the United States. Among different race groups, African-Americans are at higher risk of dying from CVD and have a worse risk factor profile. Left ventricular hypertrophy is an important risk factor in cardiovascular disease and echocardiography has been widely used for diagnosis. The data used in our first case study is belong to a subgroup of African-Americans with hypertension who are at high risk of cardiovascular/heart failure disease. Data are captured from patients admitted in the emergency department of Detroit Receiving Hospital in Detroit Medical Center (DMC). Across several attributes consisting demographic data, patient clinical records, individual health status, laboratory information, and cardiovascular magnetic resonance imaging results, 172 attributes left after preprocessing phase for data analysis related to 91 patients. As mentioned before, the goal is to predict value of heart damage risk level based on high-dimensional features.    

We implemented Deep Belief Network, Variational autoencoder and Adversarial autoencoder by using TensorFlow and Theano libraries and executed Stacked sparse autoencoder by Keras library with tensorflow backend in Python. All authoencoders except DBN are applied with 5 hidden layers (two hidden layers of encoders and decoders and one middle layer). We performed deep belief network with 3 hidden layers.

For each deep architecture, we applied parameter tuning for major parameters such as learning rate, activation functions and batch size to select the best parameters. We performed different deep networks which differ in the number of neurons in hidden layers for all autoencoders type and then select the best performance of autoencoders across all networks.

For the supervised learning step we consider three well-known supervised classifiers: Random Forests \cite{breiman2001random}, Lasso Regression \cite{tibshirani1996regression} and Support Vector Machine (SVM) \cite{suykens1999least}. We used Root Mean Squared Error (RMSE) as our evaluation measure for performance validation in testing process.

We performed our approach for different combinations of deep architectures (represented data) and supervised classifiers as well as original data (unrepresented data), and compared their performance based on the results obtained from testing process with 5-folds cross validation (for each fold we considered 80\% of the data for training, 10\% for validation and 20\% for test set), also we used the weights learned in training process to represent the data in the testing process. This comparison has been shown in Table \ref{s0}. According to this results, our approach with representation learning reduces the prediction error and achieves a better accuracy rather than using the original features. Among different combinations, using stacked sparse autoencoders for feature learning and random forests for supervised learning lead to the least RMSE for this small dataset (DMC dataset).

\begin{table}[H]
	\small 
	\centering
	\captionsetup{justification=justified, width=0.9\linewidth}
	\caption{Performance comparison among represented data and original features (DMC dataset)} \label{s0}
	
	\begin{tabular}{ c   | p{1.3 cm}  p{1.3 cm}   p{1.3 cm} }
		
		\ \textbf{Approach} &    \textbf{RF} & \textbf{Lasso} & \textbf{SVM}\\
		\hline
		\textbf{SSAE} &  \textbf{\color{blue}6.89} & \textbf{9.53} & \textbf{9.31}  \\
		\hline
		\textbf{DBN} &  7.91 & 9.81 & 10.02 \\
		\hline
		\textbf{AAE} & 8.49  & 9.89 & 10.06  \\
		\hline
		\textbf{VAE} & 9.65 & 10.17 & 9.95    \\
		\hline
		\textbf{Original } & 11.08 & 13.86 & 12.16 \\
		\hline
		
	\end{tabular}
\end{table}

\subsection{Case study 2 (Large Datasets): eICU dataset}
In the second case study, we used the eICU collaborative research database: a large, publicly available database provided by the MIT Laboratory in partnership with the Philips eICU Research Institute \cite{johnson2017analyzing}. Medical doctors predict intensive care units (ICUs) length of stay for planning ICU capacity as an expensive unit in the hospital and identifying unexpectedly long ICU length of stay in special cases to better monitoring \cite{verburg2017models}. The care provided by ICUs is complicated and the related costs are high, so ICUs are particularly interested in evaluating, planning and improving their performance \cite{verburg2014comparison}.  

The most popular approach for prediction of length of stay in ICUs is developed based on acute physiology score of APACHE (Acute Physiology and Chronic Health Evaluation) which lead to poor prediction performance in several cases \cite{verburg2017models}. APACHE introduced in 1978 for developing of severity-of-illness classification system and proposing a measure for describing different groups in ICUs and assessing their care \cite{wagner1984acute}. APACHE approaches use multivariate linear regression procedure based on acute physiology score and some other variables such as age and chronic health conditions \cite{zimmerman2006acute} to predict length of stay.

The data in the eICU database includes patients who were admitted to intensive care units during 2014 and 2015. Among different patients, we choose cardiovascular and Neurological patients admitted in the Cardiac-eICU and Neuro-eICU respectively. We integrated several features including hospital and administration data, demographics information, diagnosis and laboratory test data, drugs information, monitored invasive vital sign data and clinical patient history data. After cleaning and preprocessing step, we finalized more than 150 features for each dataset with approximately 7000 and 8000 records belonging to Cardiac-eICU and Neuro-eICU units, respectively. In this case study, we conduct the same approach as we did for the first case study and our purpose is to predict the patient length of stay (days) in these two ICU units based on high-dimensional features. 

Table \ref{s1} and \ref{s2} demonstrate RMSE results for different types of autoencoders and supervised learners in Cardiac-ICU and Neuro-ICU data respectively. Although, AAE shows a great performance, VAE's results are impressive and provides perfect prediction for length of stay. As illustrated, SSAE and DBN could not improve the model accuracy as much as AAE and VAE.    

We also considered the prediction results of APACHE approach reported in eICU collaborative research database and calculated the RMSE for both Cardiac and Neuro patients. The results indicated weak performance of APACHE approach, for instance, the RMSE for Cardiac and Neuro datasets were 8.04 and 8.12 respectively. Therefore, different machine learning approaches using original data or represented data outperformed APACHE approach significantly.    

\begin{table}[H]
	\begin{minipage}[b]{0.48\textwidth}
		\small 
		\centering
		\captionsetup{justification=justified, width=0.9\linewidth}
		\caption{Performance comparison between represented data and baseline (Cardiac-ICU)} \label{s1}		
		\begin{tabular}{ c   | p{1.3 cm}  p{1.3 cm}   p{1.3 cm} }			
			\ \textbf{Approach} &    \textbf{RF} & \textbf{Lasso} & \textbf{SVM}\\
			\hline
			\textbf{SSAE} & 1.59 & 4.08 & 2.47 \\
			\hline
			\textbf{DBN} & 0.97 & 3.78 & 2.29\\
			\hline
			\textbf{AAE} & 0.57 & 3.57 & 1.99 \\
			\hline
			\textbf{VAE} & \textbf{\color{blue}0.20} & \textbf{2.83} & \textbf{1.87}\\
			\hline
			\textbf{Original } & 1.65 & 4.14 & 2.52 \\
			\hline			
		\end{tabular}
	\end{minipage}
	\hfill
	\vspace{0.7cm}
	\begin{minipage}[b]{0.48\textwidth}
		\small 
		\centering
		\captionsetup{justification=justified, width=0.9\linewidth}
		\caption{Performance comparison between represented data and baseline (Neuro-ICU)} \label{s2}		
		\begin{tabular}{ c   | p{1.3 cm}  p{1.3 cm}   p{1.3 cm} }			
			\ \textbf{Approach} &    \textbf{RF} & \textbf{Lasso} & \textbf{SVM}\\
			\hline
			\textbf{SSAE} & 1.31 & 2.76 & 3.34 \\
			\hline
			\textbf{DBN} & 0.95 & 2.53 & 2.06\\
			\hline
			\textbf{AAE} & 0.72 & 2.28 & 2.45 \\
			\hline
			\textbf{VAE} & \textbf{\color{blue}0.14} & \textbf{2.05} & \textbf{1.94}\\
			\hline
			\textbf{Original } & 1.38 & 2.95 & 3.51 \\
			\hline			
		\end{tabular}
	\end{minipage}
	
\end{table}

\begin{figure}[h!t]
	\centering
	\subfloat[SSAE loss-DMC data ]{\includegraphics[width=0.23\textwidth]{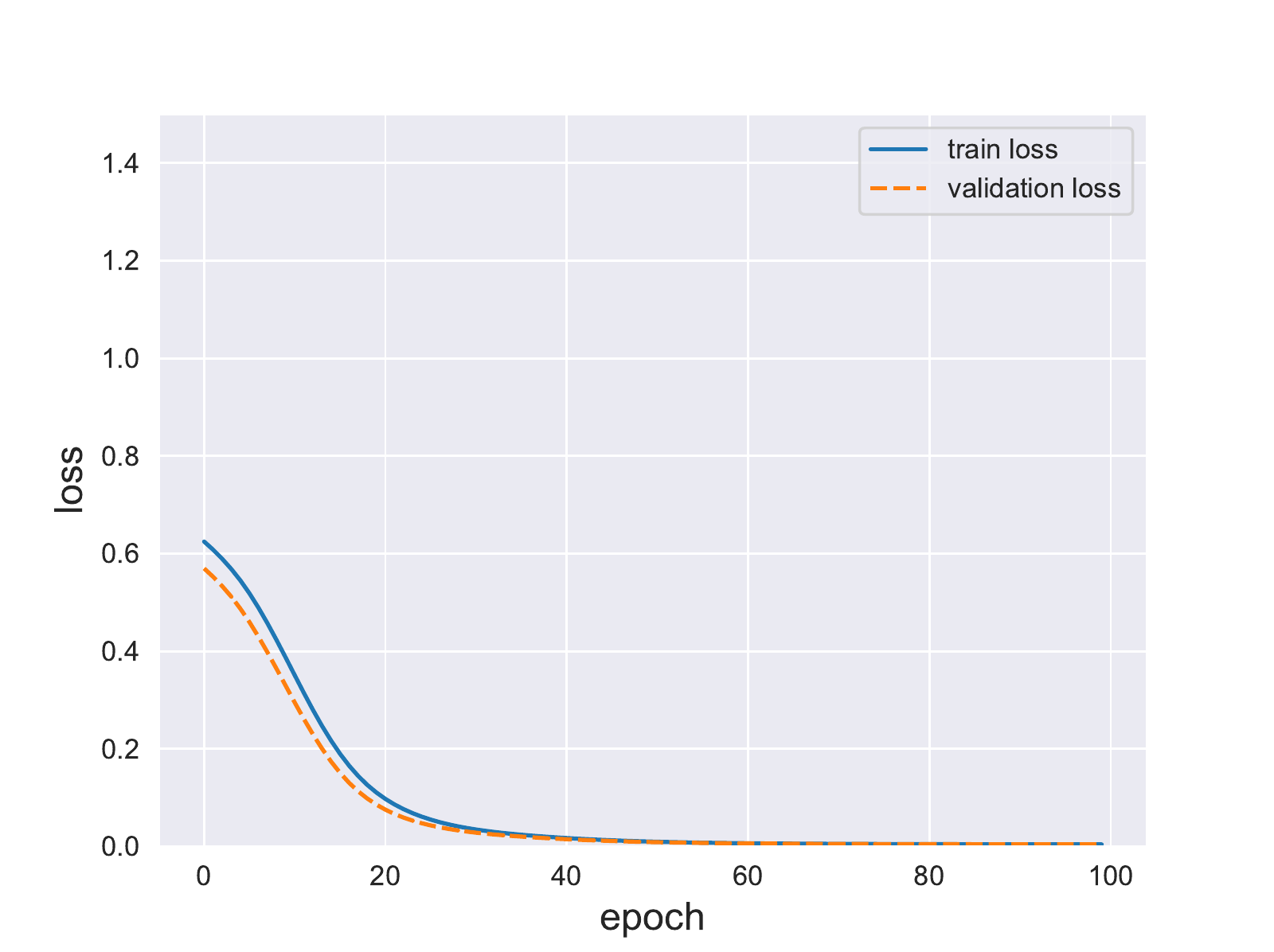}\label{fig:a}}
	\hfill
	\subfloat[SSAE loss-Cardiac ICU data ]{\includegraphics[width=0.23\textwidth]{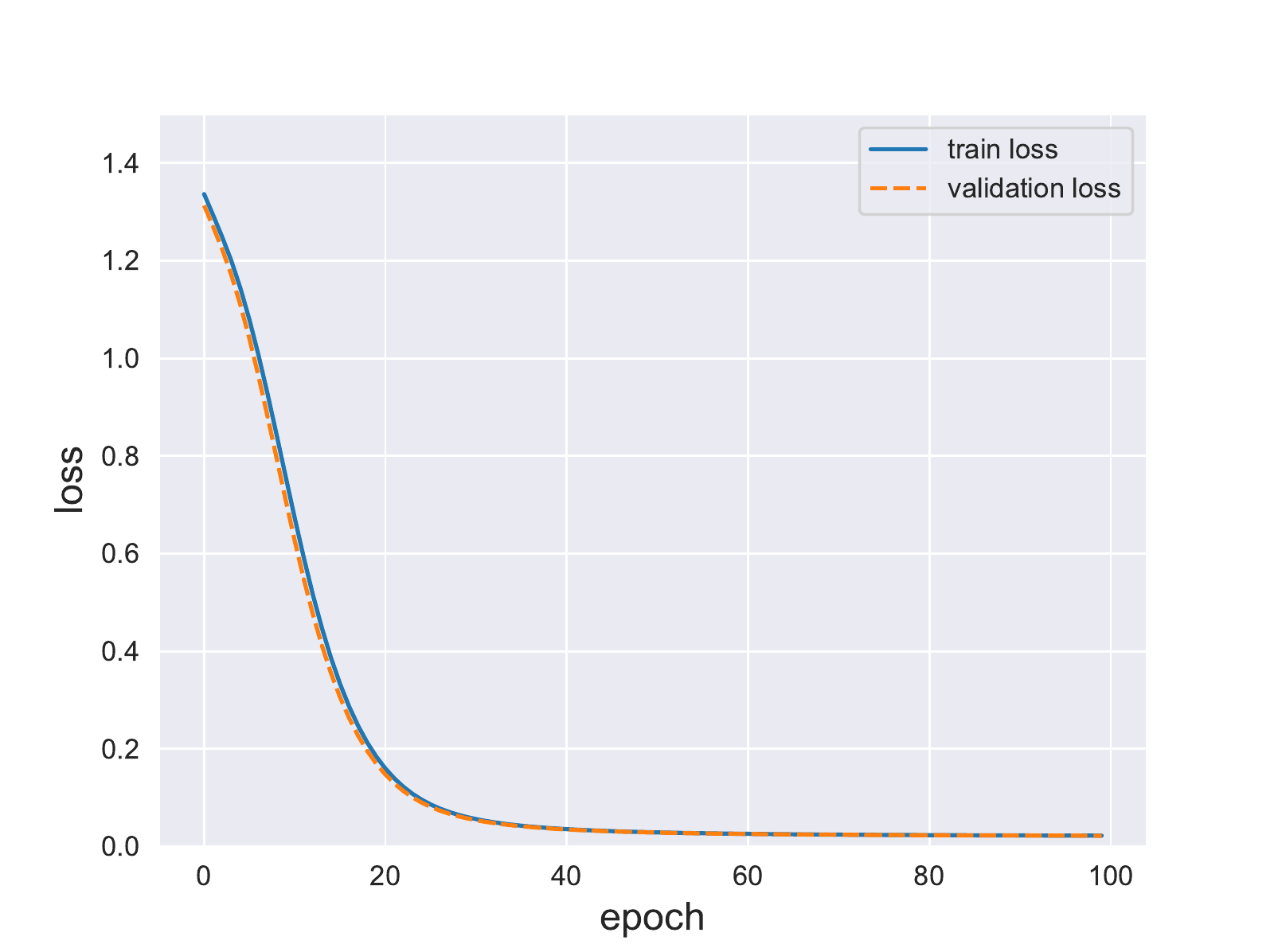}\label{fig:b}}
	\hfill
	\subfloat[VAE loss-DMC data ]{\includegraphics[width=0.23\textwidth]{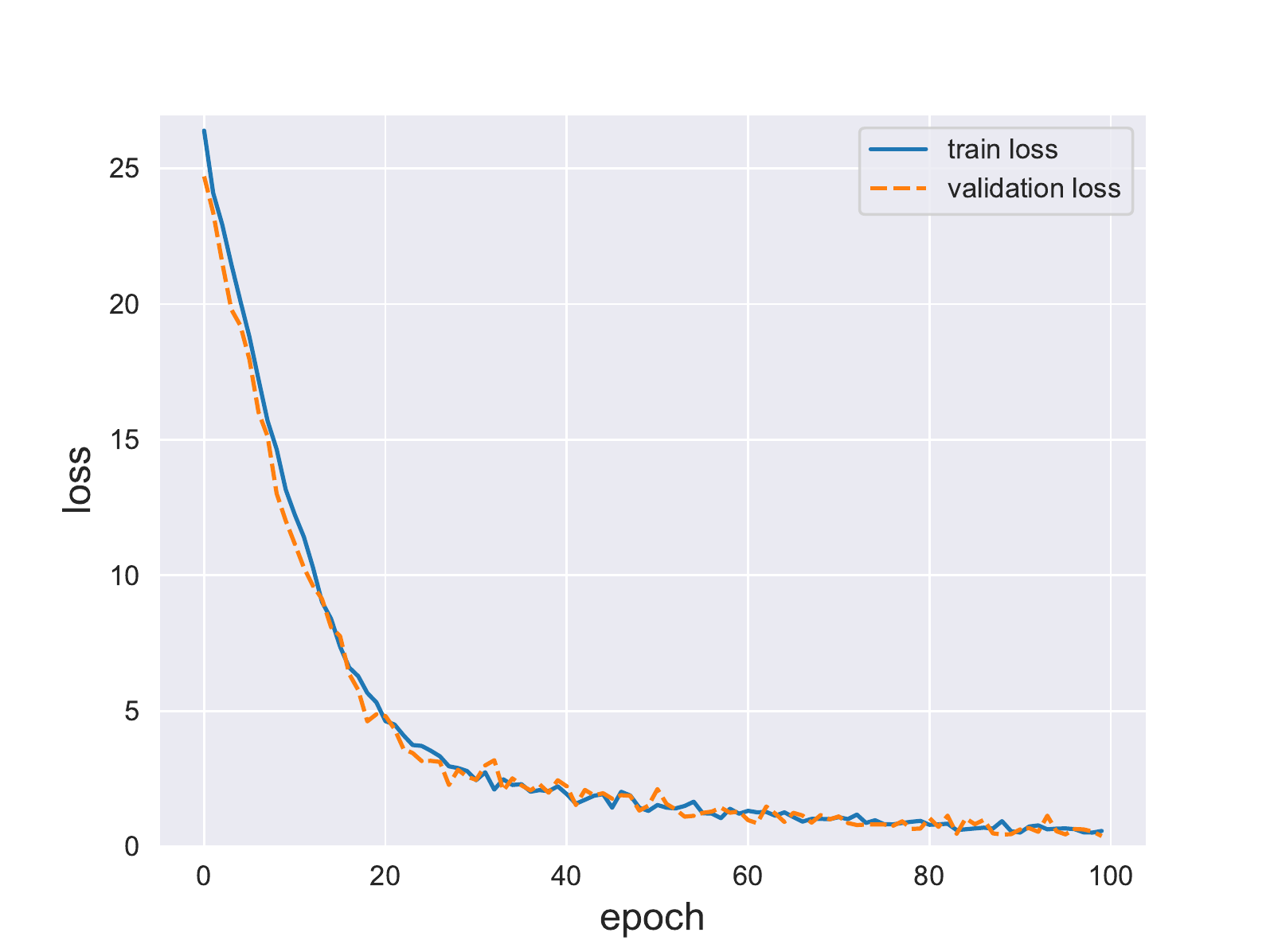}\label{fig:c}}
	\hfill
	\subfloat[VAE loss-Cardiac ICU data ]{\includegraphics[width=0.23\textwidth]{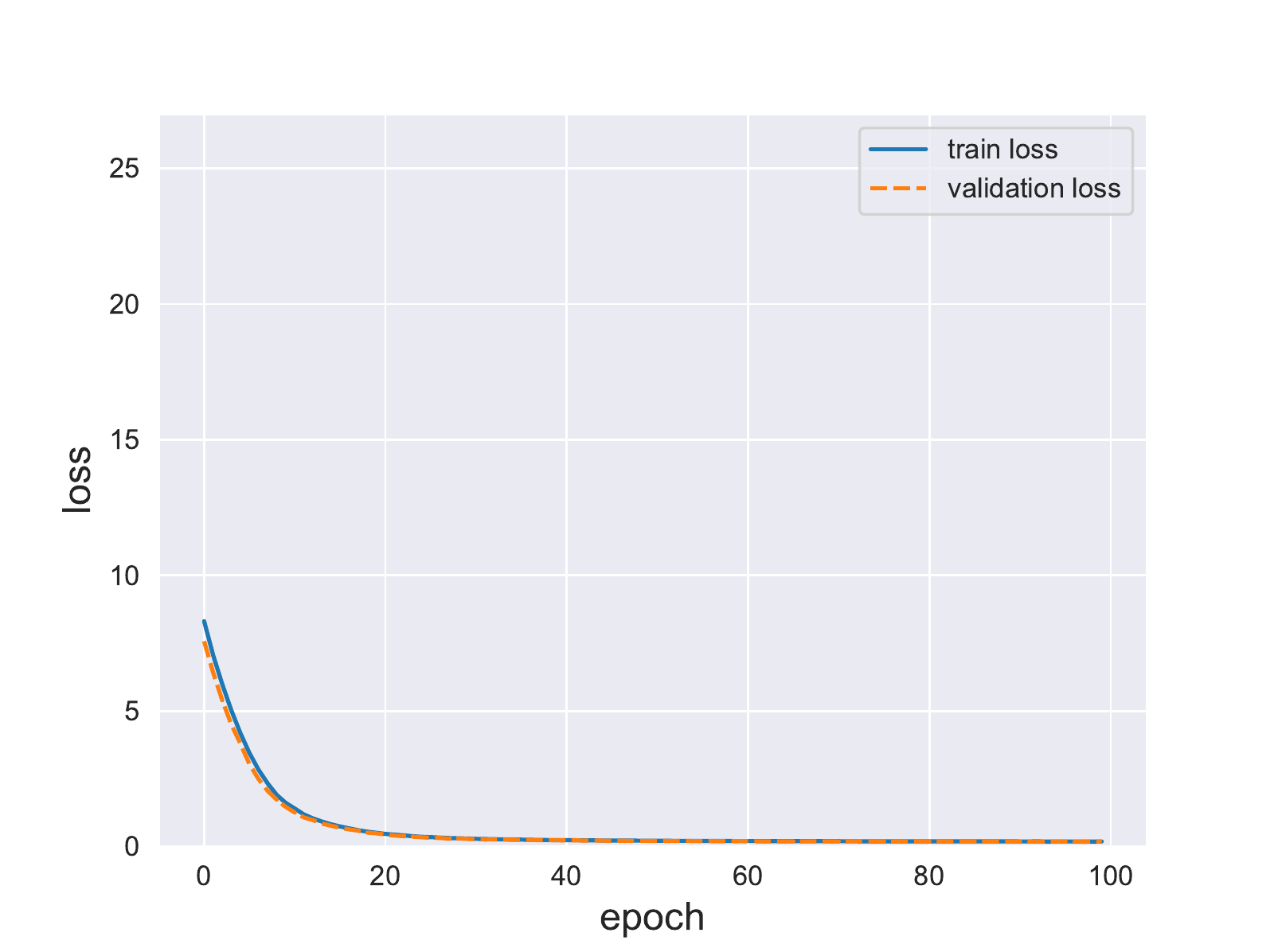}\label{fig:d}}
	\caption{Training and validation loss of SSAE and VAE in training process for small and large datasets}
	\label{DNN}
\end{figure}

To better understand SSAE and VAE performance in the representation learning, we analyze their training and validation loss for small and large datasets separately. As illustrated in Figures \ref{fig:a}, \ref{fig:b}, \ref{fig:c} and \ref{fig:d}, the training and validation loss are end up to be roughly the same and also their values are converging (good fitting). Since we used regularization in both SSAE and VAE, it leads to have less amount of loss in validation set rather than training. Based on Figures \ref{fig:a} and \ref{fig:b}, the SSAE loss (which is based on MAE loss function) in samll dataset (DMC data)  is less than SSAE loss of large dataset (Cardiac-ICU) across 100 epochs, hence the SSAE achieves better representation for small dataset. In the other side (Figures \ref{fig:c} and \ref{fig:d}), VAE loss (which is based on MSE of reconstruction error + average of KL loss) in large dataset is lower than VAE loss in small dataset, therefore provides better representation learning. 

\section{Discussion and Conclusion}
In this research, we proposed a comparative study for evaluation of deep feature representation in applications to Electronic Medical Records (EHRs). 
The results emphasize that the choice of representation learning plays an effective rule in the performance of clinical prediction. While in the first case study (small dataset), regular autoencoders (SSAE, DBN) had a better accuracy in comparison with generative autoencoders (AAE, VAE) and for large datasets (eICU database), VAE outperforms the other deep architectures significantly. In other words, we can conclude that feature representation using deep learning would be effective for both small and large datasets and choice of deep network achieves different results. The advantage of AAE and VAE in learning true distribution of input features based on distribution of sample from latent variables makes it different and it seems that these advanced generative networks achieve better representation in the case of large and more complex data in comparison with regular autoencoders such as SAE and DBN.           

Empirically, our results demonstrate that: 1) Medical feature representation can improve the performance of prediction and 2) Choice of representation can lead to different performances. This choice of representation might be related to the number of instances ($n$) and number of features ($p$) in the dataset. In future works, it is necessary to compare the performance of generative models (e.g. AAE and VAE) with some other approaches (e.g. SSAE and DBN) in different scenarios. As a guideline, we need to consider four scenarios: a) large $n$  and large $p$, b) large $n$ and small $p$, c) small $n$ and large $p$, d) small $n$ and small $p$ and generate some insights about choice of representation between advanced generative models and otherwise.

	\bibliographystyle{plain}
	\bibliography{References}{}

\begin{thebibliography}{10}

\bibitem{bair2006prediction}
Eric Bair, Trevor Hastie, Debashis Paul, and Robert Tibshirani.
\newblock Prediction by supervised principal components.
\newblock {\em Journal of the American Statistical Association},
  101(473):119--137, 2006.

\bibitem{bengio2013representation}
Yoshua Bengio, Aaron Courville, and Pascal Vincent.
\newblock Representation learning: A review and new perspectives.
\newblock {\em IEEE transactions on pattern analysis and machine intelligence},
  35(8):1798--1828, 2013.

\bibitem{bengio2007greedy}
Yoshua Bengio, Pascal Lamblin, Dan Popovici, and Hugo Larochelle.
\newblock Greedy layer-wise training of deep networks.
\newblock In {\em Advances in neural information processing systems}, pages
  153--160, 2007.

\bibitem{breiman2001random}
Leo Breiman.
\newblock Random forests.
\newblock {\em Machine learning}, 45(1):5--32, 2001.

\bibitem{cao2016deepqa}
Renzhi Cao, Debswapna Bhattacharya, Jie Hou, and Jianlin Cheng.
\newblock Deepqa: improving the estimation of single protein model quality with
  deep belief networks.
\newblock {\em BMC bioinformatics}, 17(1):495, 2016.

\bibitem{che2017boosting}
Zhengping Che, Yu~Cheng, Shuangfei Zhai, Zhaonan Sun, and Yan Liu.
\newblock Boosting deep learning risk prediction with generative adversarial
  networks for electronic health records.
\newblock In {\em Data Mining (ICDM), 2017 IEEE International Conference on},
  pages 787--792. IEEE, 2017.

\bibitem{cheng2016risk}
Yu~Cheng, Fei Wang, Ping Zhang, and Jianying Hu.
\newblock Risk prediction with electronic health records: A deep learning
  approach.
\newblock In {\em Proceedings of the 2016 SIAM International Conference on Data
  Mining}, pages 432--440. SIAM, 2016.

\bibitem{choi2016doctor}
Edward Choi, Mohammad~Taha Bahadori, Andy Schuetz, Walter~F Stewart, and Jimeng
  Sun.
\newblock Doctor ai: Predicting clinical events via recurrent neural networks.
\newblock In {\em Machine Learning for Healthcare Conference}, pages 301--318,
  2016.

\bibitem{choi2016multi}
Edward Choi, Mohammad~Taha Bahadori, Elizabeth Searles, Catherine Coffey,
  Michael Thompson, James Bost, Javier Tejedor-Sojo, and Jimeng Sun.
\newblock Multi-layer representation learning for medical concepts.
\newblock In {\em Proceedings of the 22nd ACM SIGKDD International Conference
  on Knowledge Discovery and Data Mining}, pages 1495--1504. ACM, 2016.

\bibitem{deng2019collaborative}
Xiaoyi Deng and Feifei Huangfu.
\newblock Collaborative variational deep learning for healthcare
  recommendation.
\newblock {\em IEEE Access}, 7:55679--55688, 2019.

\bibitem{doersch2016tutorial}
Carl Doersch.
\newblock Tutorial on variational autoencoders.
\newblock {\em arXiv preprint arXiv:1606.05908}, 2016.

\bibitem{goodfellow2014generative}
Ian Goodfellow, Jean Pouget-Abadie, Mehdi Mirza, Bing Xu, David Warde-Farley,
  Sherjil Ozair, Aaron Courville, and Yoshua Bengio.
\newblock Generative adversarial nets.
\newblock In {\em Advances in neural information processing systems}, pages
  2672--2680, 2014.

\bibitem{hinton2006fast}
Geoffrey~E Hinton, Simon Osindero, and Yee-Whye Teh.
\newblock A fast learning algorithm for deep belief nets.
\newblock {\em Neural computation}, 18(7):1527--1554, 2006.

\bibitem{hinton2006reducing}
Geoffrey~E Hinton and Ruslan~R Salakhutdinov.
\newblock Reducing the dimensionality of data with neural networks.
\newblock {\em science}, 313(5786):504--507, 2006.

\bibitem{johnson2017analyzing}
Alistair~EW Johnson, Tom~J Pollard, Leo~A Celi, and Roger~G Mark.
\newblock Analyzing the eicu collaborative research database.
\newblock In {\em Proceedings of the 8th ACM International Conference on
  Bioinformatics, Computational Biology, and Health Informatics}, pages
  631--631. ACM, 2017.

\bibitem{kingma2013auto}
Diederik~P Kingma and Max Welling.
\newblock Auto-encoding variational bayes.
\newblock {\em arXiv preprint arXiv:1312.6114}, 2013.

\bibitem{learning2013computer}
Deep Learning.
\newblock Computer science department.
\newblock {\em Stanford University. http://ufldl. stanford. edu/tutorial},
  20:21--22, 2013.

\bibitem{li2010gene}
Bo~Li, Chun-Hou Zheng, De-Shuang Huang, Lei Zhang, and Kyungsook Han.
\newblock Gene expression data classification using locally linear discriminant
  embedding.
\newblock {\em Computers in Biology and Medicine}, 40(10):802--810, 2010.

\bibitem{li2017sdt}
Xiangrui Li, Dongxiao Zhu, Ming Dong, Milad~Zafar Nezhad, Alexander Janke, and
  Phillip~D Levy.
\newblock Sdt: A tree method for detecting patient subgroups with personalized
  risk factors.
\newblock {\em AMIA Summits on Translational Science Proceedings}, 2017.

\bibitem{li2016deep}
Yifeng Li, Chih-Yu Chen, and Wyeth~W Wasserman.
\newblock Deep feature selection: theory and application to identify enhancers
  and promoters.
\newblock {\em Journal of Computational Biology}, 23(5):322--336, 2016.

\bibitem{litjens2017survey}
Geert Litjens, Thijs Kooi, Babak~Ehteshami Bejnordi, Arnaud Arindra~Adiyoso
  Setio, Francesco Ciompi, Mohsen Ghafoorian, Jeroen~AWM van~der Laak, Bram van
  Ginneken, and Clara~I S{\'a}nchez.
\newblock A survey on deep learning in medical image analysis.
\newblock {\em Medical image analysis}, 42:60--88, 2017.

\bibitem{ma2009identification}
Shuangge Ma and Michael~R Kosorok.
\newblock Identification of differential gene pathways with principal component
  analysis.
\newblock {\em Bioinformatics}, 25(7):882--889, 2009.

\bibitem{makhzani2015adversarial}
Alireza Makhzani, Jonathon Shlens, Navdeep Jaitly, Ian Goodfellow, and Brendan
  Frey.
\newblock Adversarial autoencoders.
\newblock {\em arXiv preprint arXiv:1511.05644}, 2015.

\bibitem{martis2012application}
Roshan~Joy Martis, U~Rajendra Acharya, KM~Mandana, Ajoy~Kumar Ray, and Chandan
  Chakraborty.
\newblock Application of principal component analysis to ecg signals for
  automated diagnosis of cardiac health.
\newblock {\em Expert Systems with Applications}, 39(14):11792--11800, 2012.

\bibitem{miotto2016deep}
Riccardo Miotto, Li~Li, Brian~A Kidd, and Joel~T Dudley.
\newblock Deep patient: An unsupervised representation to predict the future of
  patients from the electronic health records.
\newblock {\em Scientific reports}, 6:26094, 2016.

\bibitem{miotto2017deep}
Riccardo Miotto, Fei Wang, Shuang Wang, Xiaoqian Jiang, and Joel~T Dudley.
\newblock Deep learning for healthcare: review, opportunities and challenges.
\newblock {\em Briefings in Bioinformatics}, page bbx044, 2017.

\bibitem{nezhad2016safs}
Milad~Zafar Nezhad, Dongxiao Zhu, Xiangrui Li, Kai Yang, and Phillip Levy.
\newblock Safs: A deep feature selection approach for precision medicine.
\newblock In {\em Bioinformatics and Biomedicine (BIBM), 2016 IEEE
  International Conference on}, pages 501--506. IEEE, 2016.

\bibitem{ng2011sparse}
Andrew Ng et~al.
\newblock Sparse autoencoder.
\newblock {\em CS294A Lecture notes}, 72(2011):1--19, 2011.

\bibitem{nyamundanda2010probabilistic}
Gift Nyamundanda, Lorraine Brennan, and Isobel~Claire Gormley.
\newblock Probabilistic principal component analysis for metabolomic data.
\newblock {\em BMC bioinformatics}, 11(1):571, 2010.

\bibitem{pearson1901liii}
Karl Pearson.
\newblock Liii. on lines and planes of closest fit to systems of points in
  space.
\newblock {\em The London, Edinburgh, and Dublin Philosophical Magazine and
  Journal of Science}, 2(11):559--572, 1901.

\bibitem{pennington2014glove}
Jeffrey Pennington, Richard Socher, and Christopher Manning.
\newblock Glove: Global vectors for word representation.
\newblock In {\em Proceedings of the 2014 conference on empirical methods in
  natural language processing (EMNLP)}, pages 1532--1543, 2014.

\bibitem{purushotham2018benchmarking}
Sanjay Purushotham, Chuizheng Meng, Zhengping Che, and Yan Liu.
\newblock Benchmarking deep learning models on large healthcare datasets.
\newblock {\em Journal of biomedical informatics}, 83:112--134, 2018.

\bibitem{ravi2017deep}
Daniele Ravi, Charence Wong, Fani Deligianni, Melissa Berthelot, Javier
  Andreu-Perez, Benny Lo, and Guang-Zhong Yang.
\newblock Deep learning for health informatics.
\newblock {\em IEEE journal of biomedical and health informatics}, 21(1):4--21,
  2017.

\bibitem{shickel2017deep}
Benjamin Shickel, Patrick Tighe, Azra Bihorac, and Parisa Rashidi.
\newblock Deep ehr: A survey of recent advances on deep learning techniques for
  electronic health record (ehr) analysis.
\newblock {\em arXiv preprint arXiv:1706.03446}, 2017.

\bibitem{shickel2018deep}
Benjamin Shickel, Patrick~James Tighe, Azra Bihorac, and Parisa Rashidi.
\newblock Deep ehr: A survey of recent advances in deep learning techniques for
  electronic health record (ehr) analysis.
\newblock {\em IEEE journal of biomedical and health informatics},
  22(5):1589--1604, 2018.

\bibitem{suykens1999least}
Johan~AK Suykens and Joos Vandewalle.
\newblock Least squares support vector machine classifiers.
\newblock {\em Neural processing letters}, 9(3):293--300, 1999.

\bibitem{tabacof2016adversarial}
Pedro Tabacof, Julia Tavares, and Eduardo Valle.
\newblock Adversarial images for variational autoencoders.
\newblock {\em arXiv preprint arXiv:1612.00155}, 2016.

\bibitem{tenenbaum2000global}
Joshua~B Tenenbaum, Vin De~Silva, and John~C Langford.
\newblock A global geometric framework for nonlinear dimensionality reduction.
\newblock {\em science}, 290(5500):2319--2323, 2000.

\bibitem{tibshirani1996regression}
Robert Tibshirani.
\newblock Regression shrinkage and selection via the lasso.
\newblock {\em Journal of the Royal Statistical Society: Series B
  (Methodological)}, 58(1):267--288, 1996.

\bibitem{verburg2017models}
Ilona Willempje~Maria Verburg, Alireza Atashi, Saeid Eslami, Rebecca Holman,
  Ameen Abu-Hanna, Everet de~Jonge, Niels Peek, and Nicolette~Fransisca
  de~Keizer.
\newblock Which models can i use to predict adult icu length of stay? a
  systematic review.
\newblock {\em Critical care medicine}, 45(2):e222--e231, 2017.

\bibitem{verburg2014comparison}
Ilona~WM Verburg, Nicolette~F de~Keizer, Evert de~Jonge, and Niels Peek.
\newblock Comparison of regression methods for modeling intensive care length
  of stay.
\newblock {\em PloS one}, 9(10):e109684, 2014.

\bibitem{vincent2010stacked}
Pascal Vincent, Hugo Larochelle, Isabelle Lajoie, Yoshua Bengio, and
  Pierre-Antoine Manzagol.
\newblock Stacked denoising autoencoders: Learning useful representations in a
  deep network with a local denoising criterion.
\newblock {\em Journal of Machine Learning Research}, 11(Dec):3371--3408, 2010.

\bibitem{wagner1984acute}
Douglas~P Wagner and Elizabeth~A Draper.
\newblock Acute physiology and chronic health evaluation (apache ii) and
  medicare reimbursement.
\newblock {\em Health care financing review}, 1984(Suppl):91, 1984.

\bibitem{wulsin2010semi}
Drausin Wulsin, Justin Blanco, Ram Mani, and Brian Litt.
\newblock Semi-supervised anomaly detection for eeg waveforms using deep belief
  nets.
\newblock In {\em 2010 Ninth International Conference on Machine Learning and
  Applications}, pages 436--441. IEEE, 2010.

\bibitem{xiao2018opportunities}
Cao Xiao, Edward Choi, and Jimeng Sun.
\newblock Opportunities and challenges in developing deep learning models using
  electronic health records data: a systematic review.
\newblock {\em Journal of the American Medical Informatics Association}, 2018.

\bibitem{xu2009modified}
Ping Xu, Guy~N Brock, and Rudolph~S Parrish.
\newblock Modified linear discriminant analysis approaches for classification
  of high-dimensional microarray data.
\newblock {\em Computational Statistics \& Data Analysis}, 53(5):1674--1687,
  2009.

\bibitem{yao2012independent}
Fangzhou Yao, Jeff Coquery, and Kim-Anh L{\^e}~Cao.
\newblock Independent principal component analysis for biologically meaningful
  dimension reduction of large biological data sets.
\newblock {\em BMC bioinformatics}, 13(1):24, 2012.

\bibitem{yeung2001validating}
Ka~Yee Yeung, David~R. Haynor, and Walter~L. Ruzzo.
\newblock Validating clustering for gene expression data.
\newblock {\em Bioinformatics}, 17(4):309--318, 2001.

\bibitem{zhao2017learning}
Jing Zhao, Panagiotis Papapetrou, Lars Asker, and Henrik Bostr{\"o}m.
\newblock Learning from heterogeneous temporal data in electronic health
  records.
\newblock {\em Journal of biomedical informatics}, 65:105--119, 2017.

\bibitem{zhao2016multiscale}
Liya Zhao and Kebin Jia.
\newblock Multiscale cnns for brain tumor segmentation and diagnosis.
\newblock {\em Computational and mathematical methods in medicine}, 2016, 2016.

\bibitem{zhong2019shallow}
Guoqiang Zhong, Xiao Ling, and Li-Na Wang.
\newblock From shallow feature learning to deep learning: Benefits from the
  width and depth of deep architectures.
\newblock {\em Wiley Interdisciplinary Reviews: Data Mining and Knowledge
  Discovery}, 9(1):e1255, 2019.

\bibitem{zimmerman2006acute}
Jack~E Zimmerman, Andrew~A Kramer, Douglas~S McNair, and Fern~M Malila.
\newblock Acute physiology and chronic health evaluation (apache) iv: hospital
  mortality assessment for today’s critically ill patients.
\newblock {\em Critical care medicine}, 34(5):1297--1310, 2006.

\bibitem{zou2006sparse}
Hui Zou, Trevor Hastie, and Robert Tibshirani.
\newblock Sparse principal component analysis.
\newblock {\em Journal of computational and graphical statistics},
  15(2):265--286, 2006.

\end{thebibliography}

\end{document}